\documentclass[letterpaper, 10 pt, conference]{ieeeconf} 
\IEEEoverridecommandlockouts  
\usepackage{graphics} 
\usepackage{epsfig} 
\usepackage{mathptmx} 
\usepackage{times} 
\usepackage{amsmath} 
\usepackage{amssymb}  
\usepackage{multirow}
\usepackage{diagbox}
\usepackage{xpatch}
\usepackage{hyperref}
\hypersetup{%
    pdfborder = {0 0 0}
}

\title{\LARGE \bf
TRTM: Template-based Reconstruction and Target-oriented Manipulation of Crumpled Cloths
}

\author{Wenbo Wang, Gen Li, Miguel Zamora, and Stelian Coros
\thanks{The authors are with ETH Zurich, Switzerland. \{wenbwang\}@student.ethz.ch, \{gen.li, mimora, scoros\}@inf.ethz.ch}
}

\begin{document}

\maketitle
\thispagestyle{empty}
\pagestyle{empty}

\begin{abstract}


Precise reconstruction and manipulation of the crumpled cloths is challenging due to the high dimensionality of cloth models, as well as the limited observation at self-occluded regions. We leverage the recent progress in the field of single-view reconstruction to template-based reconstruct the crumpled cloths from their top-view depth observations only, with our proposed sim-real registration protocols. In contrast to previous implicit cloth representations, our reconstruction mesh explicitly describes the positions and visibilities of the entire cloth mesh vertices, enabling more efficient dual-arm and single-arm target-oriented manipulations. Experiments demonstrate that our TRTM system can be applied to daily cloths that have similar topologies as our template mesh, but with different shapes, sizes, patterns, and physical properties. Videos, datasets, pre-trained models, and code can be downloaded from our project website: \url{https://wenbwa.github.io/TRTM/}.

\end{abstract}

\section{Introduction}


Cloth products have been a crucial part of our daily life, where repeated human resources are spent on cloth arranging tasks. For this reason, several studies have been performed to identify \cite{identify1}, perceive \cite{pixel1}, and organize \cite{organize1, organize2} different cloth items using both computer vision and robotic approaches.

However, manipulating while perceiving the entire state of one crumpled cloth is challenging due to the complex cloth model and the limited observation at self-occluded regions. In previous research, crumpled cloths are mostly represented as either visible pixel values \cite{pixel2, clothfeature4, SpeedFolding}, sampled surface points \cite{VCD}, sparse feature groups \cite{clothfeature1, clothfeature2}, or encoded latent vectors \cite{dynamics1latent}, as shown in Fig. \ref{Figure1}. They train or optimize their implicit manipulation policies with those implicit and simplified cloth representations by either reinforcement learning or dynamics learning mostly within the simulation environment. Few of the previous work fully and precisely understand the entire crumpled cloth configuration, not to mention explicitly locating and manipulating the visible cloth mesh vertices.

Different from the previous studies, we employ the template-based graph neural networks (GNNs) to explicitly reconstruct the entire meshes of crumpled cloths, using their top-view depth observations only. We demonstrate that, with our sim-real registration protocols, the distribution of cloth configurations in the real world can be properly described using adequate simulated cloth meshes, among which the ground truth mesh of each depth observation is known. Benefiting from our explicit mesh representation, the task of manipulating crumpled cloths to some target configurations can be more efficiently performed with fewer operation episodes. Our work contributes to the following:


\begin{figure}[t]
  \centering
  \includegraphics[width=0.48\textwidth]{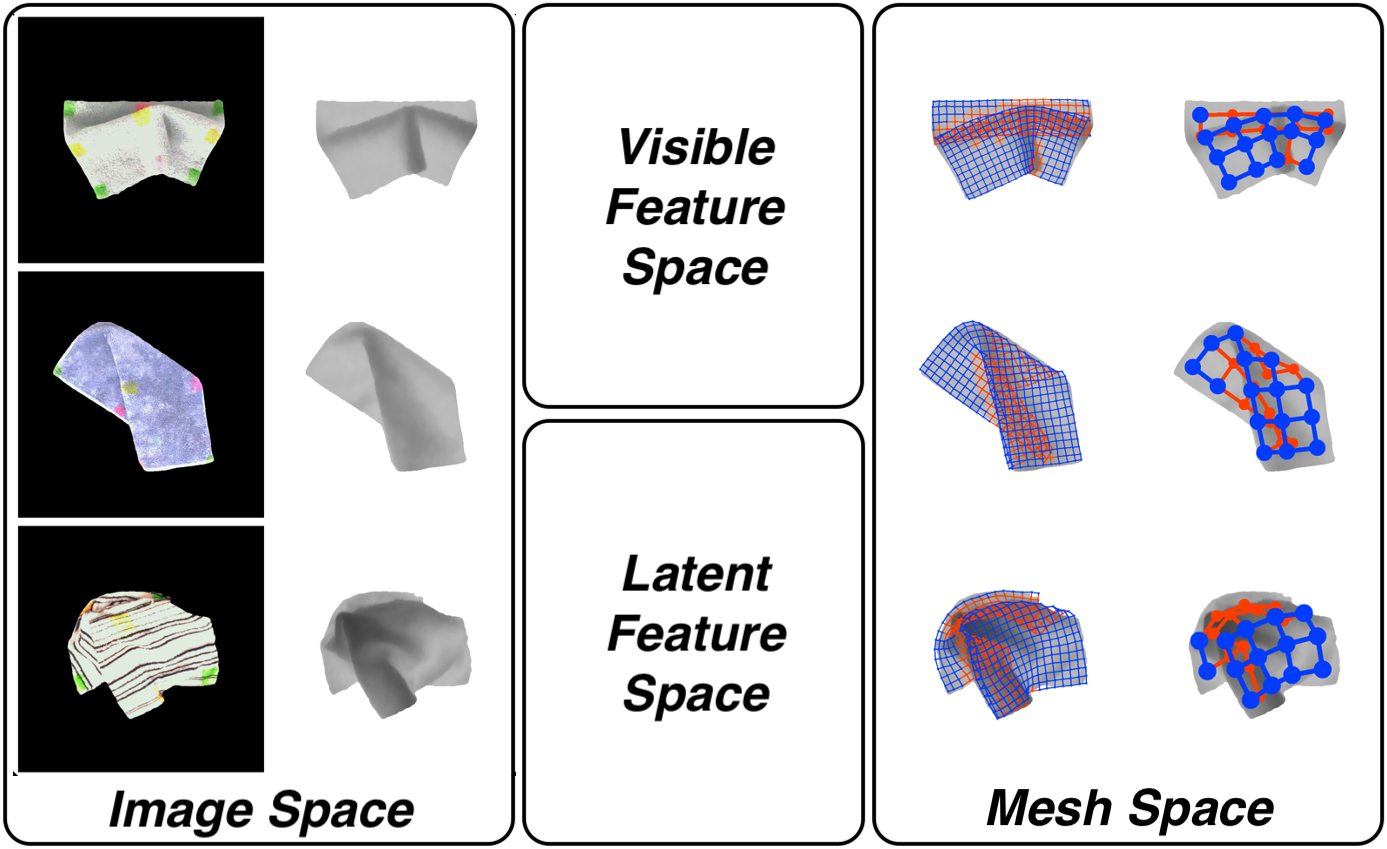}
  \caption{Different state representations of crumpled cloths. From left to right: top-view color images; top-view depth images or point clouds; sparse visible features or encoded latent vectors; our template-based reconstruction mesh and clustered mesh group for robot manipulation. From top to bottom: configurations of the randomly one-time dragged rectangle cloth, two-times folded template square cloth, and one-time dropped larger square cloth.}
  \label{Figure1}
\end{figure}

1) a novel template-based reconstruction method that can explicitly predict the positions and visibilities of the entire cloth mesh vertices from its top-view depth observation only.

2) a synthetic dataset with 120k+ simulated cloth meshes and rendered top-view RGBD images, together with one real-world dataset consisting of 3k+ collected cloth configurations and keypoint-labeled top-view RGBD images.

3) a robot system that can manipulate crumpled cloths to some target or near-target configurations by querying and selecting corresponding visible mesh vertices.

\section{Related Work}

Research on the cloth perception and manipulation is extensive. Earlier approaches typically employ handcrafted or learning-based methods to identify specific cloth features, such as corners, edges, and wrinkles, within the top-view color or depth images \cite{clothfeature1, clothfeature2, clothfeature3}. Their manipulation policies are mostly generated from those independently detected image features \cite{towels, garments}, which may be noisy, sparse, sensitive, and ambiguous. As shown in Fig. \ref{Figure1}, it is intrinsically difficult to pixel-wise distinguish all those visible cloth corners, edges, and wrinkles.

Other studies try to simplify the infinite configuration space by firstly hanging the crumpled cloth in midair, from which some feature detection \cite{hangfeature2}, mesh matching \cite{hangmesh1, hangmesh2, GarmentNet}, and dual-arm manipulation  \cite{hangfeature1} can be more easily performed. In contrast to these approaches, we focus on directly reconstructing and manipulating the initially crumpled cloths, using their top-view observations only.

\begin{figure*}[t]
  \centering
  \includegraphics[width=0.8\paperwidth]{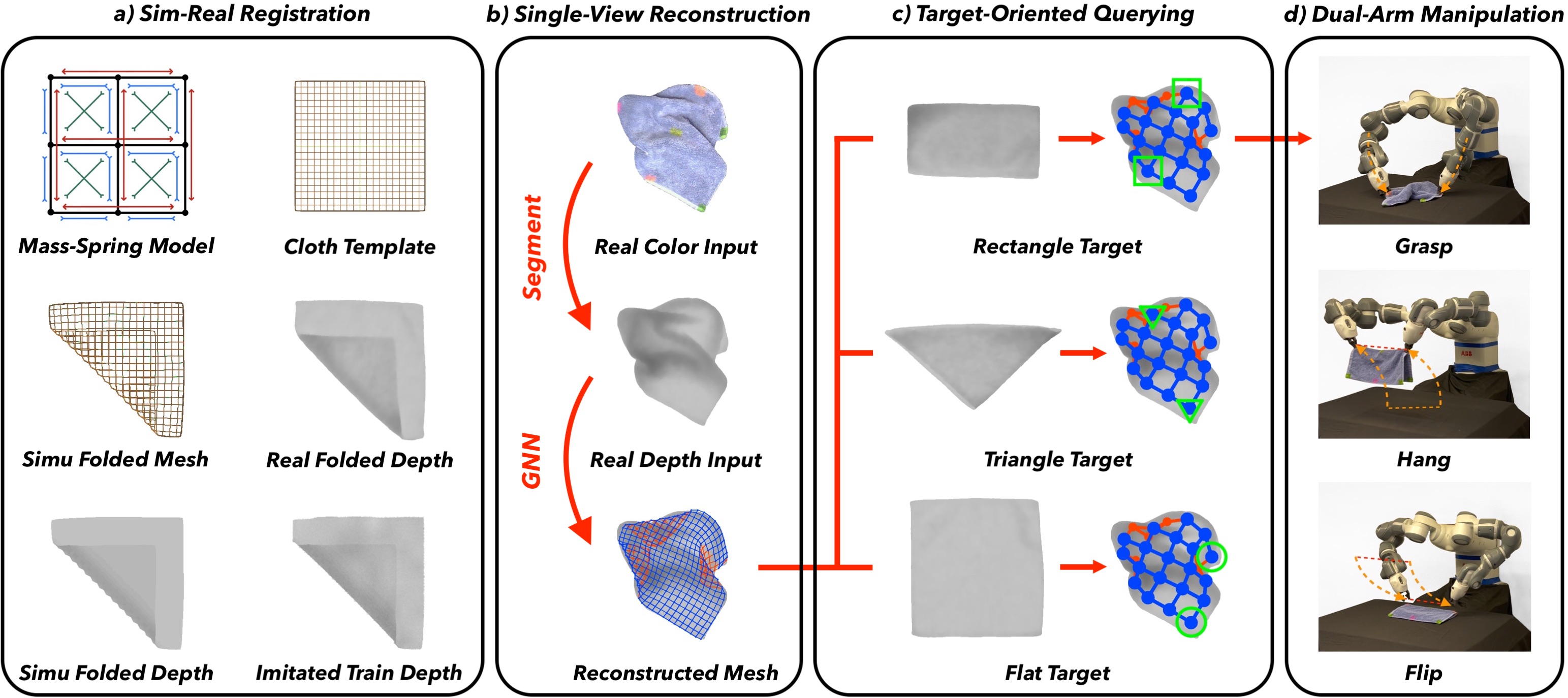}
  \caption{{\bf System Overview.} a) Sim-real registration of one real-world cloth to a synthetic mass-spring cloth mesh with imitated top-view depth observations. b) Single-view template-based reconstruction of a crumpled cloth from its top-view depth observation only, using our template-based cloth GNN. c) Querying the best visible vertex pairs within the reconstructed and clustered mesh group, according to different target configurations: flat, triangle, and rectangle. d) Dual-arm manipulation using one ABB YuMi Robot at the selected cloth vertex pair with optimized grasp-hang-and-flip trajectories.}
  \label{Figure2}
\end{figure*}

Recently, data-driven methods have been proposed to achieve more sophisticated perception and manipulation of crumpled cloths. In these studies, some parameterized single-arm or dual-arm actions, together with some task-specific folding and unfolding policies, are trained in the ways of reinforcement learning with image-based rewards \cite{reinforce1pixel, reinforce2pixel}, deep imitation learning from predefined policies \cite{imitate, organize3imitation}, or value function learning at pixel observations \cite{SpeedFolding, FlingBot}, during which the cloth deformations are not explicitly perceived. In addition to those model-free learning methods, some other studies directly optimize the random-shooting actions by dynamics learning within the simulation environment \cite{VCD, garmentdynamic, imagedynamics}, where the synthetic cloth may look and perform differently from the real-world cloths that have various textures and physical properties. In summary, few of the above work fully understands the crumpled cloth configuration, not to mention explicitly locating the visible cloth vertices and manipulating them with some target configurations.

Inspired by the above work and the recent progress in the learning-based reconstruction \cite{humanGNN}, we aim to achieve precise mesh reconstruction of the randomly dragged, folded, and dropped cloths from their top-view depth observations only. Compared with the previous implicit and simplified cloth representations, our reconstruction mesh explicitly indicates the entire cloth mesh vertices' positions and visibilities, as shown in Fig. \ref{Figure1}. Experiments demonstrate that our explicit mesh representation promotes more explicit dual-arm and single-arm target-oriented manipulations, which are more efficient and more similar to the human-wise decision, i.e., in front of a real-world cloth, we mostly generate actions from its 3D mesh embedding, instead of those 2D pixel observations, latent feature vectors, or random shooting optimizations, that are used by the previous work.

\section{Methodology}

\subsection{Sim-Real Registration}

Different from the previous studies that mostly train their policies within simulation and employ the sim2real transfer either directly \cite{VCD, sim2real} or rely on fine-tuning \cite{FlingBot, garmentdynamic}, we perform several sim-real registrations to directly shrink the gap between the simulation and the real world.

{\bf Cloth model registration}. Concretely, we register one $ 0.3m \times 0.3m $ real-world cloth used in VCD \cite{VCD} to one mass-spring cloth mesh within Blender \cite{Blender}, a physical engine that provides powerful simulation tools. The registered synthetic cloth mesh has $ 21 \times 21 $ vertices, and its simulation parameters, like bending stiffness and thickness, are manually tuned from the real-world depth observations of the folded wrinkle size and thick difference, as shown in Fig. \ref{Figure2}. 

{\bf Depth observation registration}. Both in simulation and the real world, we centralize each cloth mesh around its image center and normalize the depth observation with a constant scale. Without loss of generality, the cloth image region is one-time scaled according to the longest canonical edge length with a constant ratio of $ l_{cloth} : l_{image} = 2 : 3 $, which generalizes our reconstruction model to other cloths with different shapes and sizes, as shown in the Ablation Study. Finally, we introduce Gaussian noises to imitate camera noise and non-smooth cloth surfaces observed in the real world. These sim-real registrations only need to be performed once when generating the synthetic dataset and training the GNN.



\begin{figure*}[t]
  \centering
  \includegraphics[width=0.8\paperwidth]{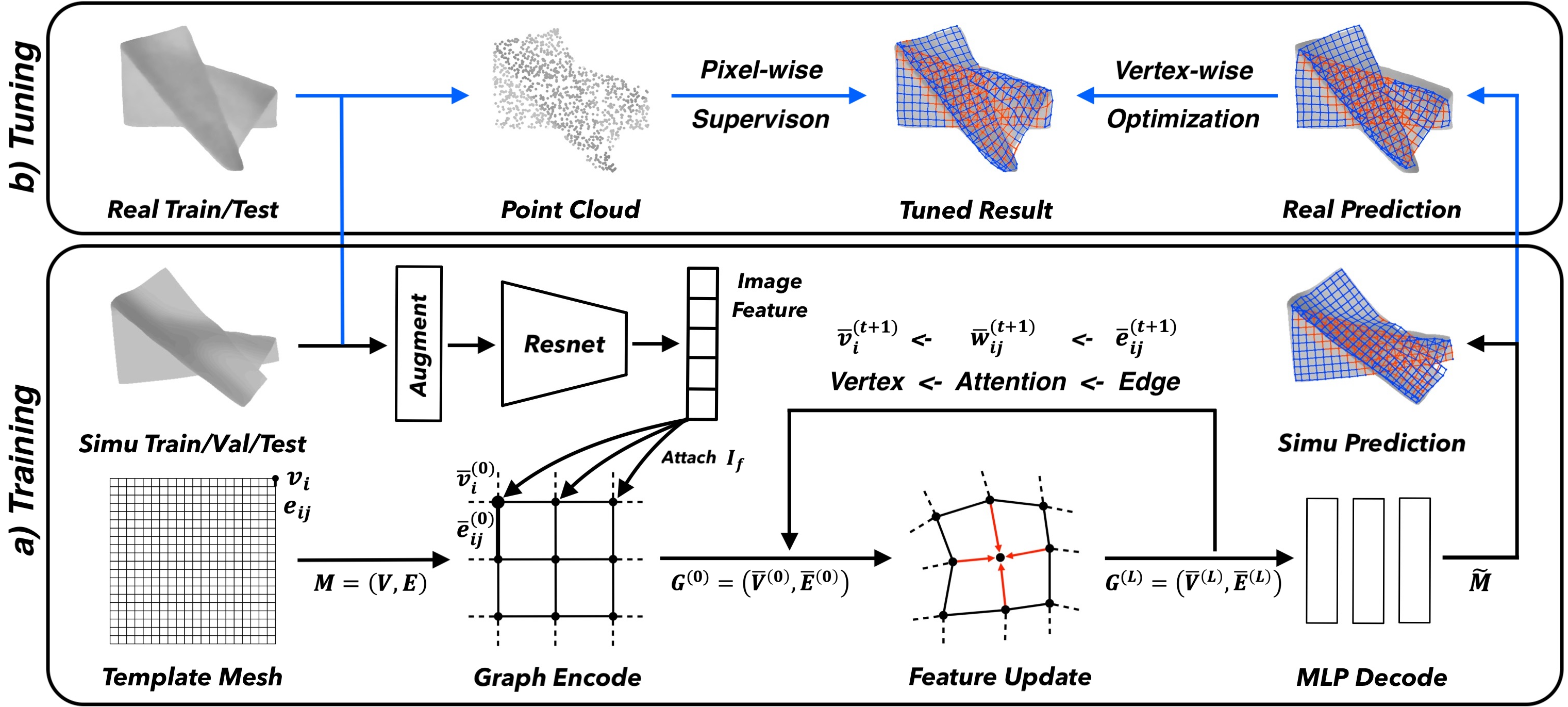}
  \caption{{\bf Template-based GNN.} a) Synthetic training with simulated cloth meshes and depth images. From left to right: synthetic cloth dataset and template mesh, image feature encoding and template graph encoding, graph feature updating with attention message flow, mesh decoding and supervising. b) Real-world tuning with collected cloth configurations and depth images. From left to right: real-world cloth dataset, point cloud observation, pixel-wise tuned result from the GNN prediction. In our work, we observe small improvements during the tuning process, as discussed in the Ablation Study.}
  \label{Figure3}
\end{figure*}

\subsection{Single-view Reconstruction}


We employ the template-based GNN from the single-view human body reconstruction work \cite{humanGNN} to our crumpled cloth setting. We design our mass-spring cloth GNN as the encoder, updater, and decoder, as described below.

{\bf Mass-spring Cloth Model.} As shown in Fig. \ref{Figure3}, one cloth can be represented as a mass-spring mesh $ M =(V, \; E) $ \cite{MassSpring} which contains a vertex group $ V = \left\{ v_{i} \right\} $ and a bidirectional edge group $ E = \left\{ e_{ij} \right\} $, where $ i,\;j=1:N^{v} $. Each mesh vertex $ v_{i} $ here contains a 3D position vector $ p_{i} $ and a visible flag $ f_{i} $ inferred from its surrounding vertices. Each mesh edge $ e_{ij} $ here connects two neighboring vertices and consists of their relative position $ p_{j}-p_{i} $ and length $ || e_{ij} ||_{2} = || p_{j}-p_{i} ||_{2} $.

{\bf Template Graph Encoding.} Our GNN encoder includes the image feature encoding and the template graph encoding. One ResNet is used here as a backbone feature extractor to encode the depth observation $ I $ into one image feature $ I_{f} $. This image feature, together with the vertices $ v_{i} $ and edges $ e_{i} $ of our template mesh $ M $, are encoded into a template graph $ G = (\bar{V}, \; \bar{E}) $ which contains one vertex feature group $ \bar{V} = \left\{ \bar{v}_{i} \right\}_{i=1:N^{v}} $ and one edge feature group $ \bar{E} = \left\{ \bar{e}_{ij} \right\}_{i,j=1:N^{v}} $:

$$
\bar{v}_{i} = MLP_{V}([p_{i}, \; I_{f}]), \; I_{f} = ResNet(I) \eqno{(1)}
$$

$$
\bar{e}_{ij} = MLP_{E}([p_{j}-p_{i}, \; \left \| p_{j}-p_{i} \right \|_{2}]) \eqno{(2)}
$$

The encoded template graph $ G^{(0)} = ({\bar{V}^{(0)}}, \; {\bar{E}^{(0)}}) $ will be iteratively updated through the attention message flow and finally decoded into the predicted cloth mesh $ \tilde{M} =(\tilde{V}, \; \tilde{E}) $.

{\bf Graph Attention Updating.} In the vanilla GNN, edge features $ \bar{e}_{ij}^{(t)} $ and vertex features $ \bar{v}_{i}^{(t)} $ are updated iteratively by averaging their neighboring features. To further improve this message updating efficiency, we introduce the attention mechanism to update vertex features by pooling neighboring edge features with learnable attention weights, following the GAT work \cite{GAT}. During the multi-step feature regression, the neural network will learn on its own which edge connection is more informative, with the weight $ \bar{w}_{ij}^{(t+1)} $ shown below:

$$
\bar{e}_{ij}^{(t+1)} = \phi_{E}^{(t)}([\bar{e}_{ij}^{(t)}, \; \bar{v}_{i}^{(t)}, \; \bar{v}_{j}^{(t)}]) \eqno{(3)}
$$

$$
\bar{w}_{ij}^{(t+1)} = \frac{exp(\phi_{A}^{(t)}([\bar{e}_{ij}^{(t+1)}]))}{\sum exp(\phi_{A}^{(t)}([\bar{e}_{ik}^{(t+1)}]))}, \; k\in Neighbor(i) \eqno{(4)}
$$

$$
\bar{v}_{i}^{(t+1)} = \phi_{V}^{(t)}([\bar{v}_{i}^{(t)}, \; \sum \bar{w}_{ik}^{(t+1)} \: \bar{e}_{ik}^{(t+1)}]), \; k\in Neighbor(i) \eqno{(5)}
$$

In our work, the edge $ \phi_{E}^{(t)} $, attention $ \phi_{A}^{(t)} $, and vertex $ \phi_{V}^{(t)} $ updaters at each iteration are parameterized using MLPs. The contribution of our learnable attention weights is around $11\%$ vertex-wise loss decay, as shown in the Ablation Study.

{\bf Cloth Mesh Decoding.} After $ L=15 $ times of the above graph updating, we decode each vertex feature $ \bar{v}_{i}^{(L)} $ into the predicted vertex position $ {\tilde{p}}_{i} $, which can be supervised within our synthetic dataset. The visible flag $ {\tilde{f}}_{i} $ at each vertex is determined by checking whether it is on the top layer of the cloth mesh within a vertical cylinder voxel: $ Voxel({\tilde{p}}_{i}) $.

$$
{\tilde{p}}_{i} = MLP_{D}([\bar{v}_{i}^{(L)}]) \eqno{(6)}
$$

$$
{\tilde{f}}_{i} = TOP_{L}({\tilde{p}}_{i}, \; {\tilde{p}}_{k}), \; {\tilde{p}}_{k} \in Voxel({\tilde{p}}_{i}) \eqno{(7)}
$$

In our work, we visualize visible vertices with blue and hidden vertices with red, while targeting to explicitly locate and manipulate different visible mesh vertices.

{\bf Implementation Details.} The above template-based cloth GNN is supervised by the randomly dragged, folded, and dropped cloth meshes simulated within Blender, together with their rendered top-view depth images. The synthetic training loss consists of five terms, as described below:

$$
L_{train} = L_{vtx, p} + \lambda_{k}L_{key, p} + \lambda_{s}L_{sil} + \lambda_{c}L_{cham} + \lambda_{r}L_{regu} \eqno{(8)}
$$

The first loss term $ L_{vtx, p} $ is the L1 distance between the predicted $ \tilde{p}_{i} $ and the ground truth vertex positions $ \hat{p}_{i} $. The second loss term $ L_{key, p} $ is one additional L1 loss at nine cloth keypoints: four corners, four middle points of edges, and one center. These two vertex-wise losses work together to assign the vertex corresponding between the predicted $ \tilde{M} $ and the ground truth cloth mesh $ \hat{M} $, as shown below:

$$
L_{vtx, p} = \frac{1}{N^{v}} \: \sum || \tilde{p}_{i} - \hat{p}_{i} \; ||_{1}, \; i=1:{N^{v}} \eqno{(9)}
$$

$$
L_{key, p} = \frac{1}{9} \: \sum || \tilde{p}_{i} - \hat{p}_{i} \; ||_{1}, \; i \in Keypoints \eqno{(10)}
$$

The third loss term $ L_{sil} $ is the image difference between the ground truth $ \hat{S} $ and the predicted silhouette $ \tilde{S} $ rendered from a differentiable renderer Pytorch3D \cite{Pytorch3D}. The fourth term $ L_{cham} $ is the unidirectional chamfer loss from the observed depth point cloud $ \hat{D} $ to the predicted vertex positions $ \tilde{P} $ \cite{garmentdynamic}. These two self-supervised pixel-wise losses work together to refine the boundary and surface similarities between the predicted $ \tilde{M} $ and the ground truth mesh $ \hat{M} $. The last term $ L_{regu} $ is a regularization loss for edge lengths.

$$
L_{sil} = \frac{1}{||\:\hat{S}\:||_{2}^{2}} \: \sum_{p\in \hat{S}} \; || \tilde{S}_{p} - \hat{S}_{p} ||_{2}^{2} \eqno{(11)}
$$

$$
L_{cham} = \frac{1}{|\hat{D}|} \: \sum_{\hat{d}_{i}\in \hat{D}} \; \min_{\tilde{p}_{k}\in \tilde{P}} \: || \tilde{p}_{k} - \hat{d}_{i} ||_{2}^{2} \eqno{(12)}
$$
\vspace{-0.4em}
$$
L_{regu} = \frac{1}{N^{e}} \: \sum | || \tilde{e}_{ij} ||_{2} - || \hat{e}_{ij} \; ||_{2}|, \; i, j=1:{N^{v}} \eqno{(13)}
$$

We set the above loss ratios as $ \lambda_{k}=1 $, $ \lambda_{s}=0.5 $, $ \lambda_{c}=0.5 $, and $ \lambda_{r}=1 $ respectively. During the synthetic training, we augment the dataset by introducing Gaussian noises and rotating with random angles. Inspired by the previous study \cite{FlingBot}, we augment each test observation by rotating eight times while predicting their cloth meshes, among which the best mesh prediction is selected using the self-supervised pixel-wise losses. The contributions of these augmentation steps are demonstrated in the Ablation Study. 

The synthetic training process takes 2 days on one 2080Ti GPU, while the inference time is around 30ms per image.

\subsection{Target-oriented Querying Policy}

Unlike the previous implicit manipulation work \cite{SpeedFolding, FlingBot}, we use our template-based reconstruction mesh to generate our dual-arm grasp-hang-and-flip actions as shown in Fig. \ref{Figure2}. In principle, each visible mesh vertex can be queried and selected for real-world robot manipulation. However, due to the non-negligible gripper size in practice and the prediction uncertainty at individual vertices, we cluster the reconstruction mesh $ \tilde{M} =(\tilde{V}, \; \tilde{E}) $ into a lower-dimensional mesh group $ \tilde{M}^{g} = (\tilde{V}^{g}, \; \tilde{E}^{g}) $ to make our querying policy more efficient and more robust, as shown in Fig. \ref{Figure1} and \ref{Figure4}. 

The clustered mesh group explicitly indicates the positions and visibilities of different cloth regions, from which we can explicitly dual-arm flip a crumpled cloth to some target or near-target configurations. To do so, we dual-arm flip the real-world cloth through each pair of the group vertices. For each target configuration, such as flat, triangle, and rectangle, we score the group vertex pairs by evaluating the silhouette difference between the flipped and the targeted cloth configurations, from which a hierarchical querying list can be generated. At each operation time, the manipulation agent will search through the above querying list within the clustered mesh group and report the best visible group vertex pair $ ({\tilde{p}}_{L}^{g_{real}}, \; {\tilde{p}}_{R}^{g_{real}}) $ for the real-world robot manipulation.


\subsection{Dual-arm Robotic Manipulation}

We formulate our dual-arm grasp-hang-and-flip actions according to the FlingBot work \cite{FlingBot}, as shown in Fig. \ref{Figure2}. One ABB YuMi robot is used here to achieve the manipulation task, during which the joint trajectories are optimized from the gripping trajectories using Newton's method \cite{ABBcontrol}. We invite interested readers to check the supplementary video and our project website for manipulation demos.


\renewcommand{\arraystretch}{1.4}
\begin{table}[t]
    \centering
    \caption{Quantitative Evaluation of Template-based Reconstruction.}
    \begin{tabular}{|c||cccccc|}
    \hline
    \multirow{2}{*}{} &
      \multicolumn{6}{c|}{Averaged over Dragged, Folded, Dropped Cloths} \\ \cline{2-7} 
     &
      \multicolumn{1}{c|}{\bf $ {T}_{simu} $} &
      \multicolumn{1}{c|}{\bf $ {T}_{real} $} &
      \multicolumn{1}{c|}{$ {Sq}_{s} $} &
      \multicolumn{1}{c|}{$ {Sq}_{l} $} &
      \multicolumn{1}{c|}{$ Rect $} &
      $ Shirt $ \\ \hline \hline
    
    $ {L}_{vtx,f} (\%) $ &
      \multicolumn{1}{c|}{10.5} &
      \multicolumn{5}{c|}{\multirow{2}{*}{None}}
      \\ \cline{1-2}
    
    $ {L}_{vtx,p} (mm) $ &
      \multicolumn{1}{c|}{{\bf 11.8}} &
      \multicolumn{5}{c|}{} \\ \hline
    
    $ {L}_{key,f} (\%) $ &
      \multicolumn{1}{c|}{11.6} &
      \multicolumn{1}{c|}{14.5} &
      \multicolumn{1}{c|}{15.0} &
      \multicolumn{1}{c|}{16.4} &
      \multicolumn{1}{c|}{18.9} &
      22.3 \\ \hline
      
    $ {L}_{key,p} (mm) $ &
      \multicolumn{1}{c|}{{\bf 12.6}} &
      \multicolumn{1}{c|}{{\bf 17.3}} &
      \multicolumn{1}{c|}{15.8} &
      \multicolumn{1}{c|}{21.5} &
      \multicolumn{1}{c|}{20.9} &
      26.7 \\ \hline
    
    $ {L}_{sil} (\%) $ &
      \multicolumn{1}{c|}{5.8} &
      \multicolumn{1}{c|}{8.1} &
      \multicolumn{1}{c|}{7.6} &
      \multicolumn{1}{c|}{8.7} &
      \multicolumn{1}{c|}{9.2} &
      11.3 \\ \hline
    
    $ {L}_{cham} (mm) $ &
      \multicolumn{1}{c|}{7.3} &
      \multicolumn{1}{c|}{9.6} &
      \multicolumn{1}{c|}{7.8} &
      \multicolumn{1}{c|}{11.5} &
      \multicolumn{1}{c|}{10.4} &
      14.9 \\ \hline \hline

      $\bf  {T}_{loss} $ &
      \multicolumn{1}{c|}{\bf 19.2} &
      \multicolumn{1}{c|}{\bf 26.1} &
      \multicolumn{1}{c|}{\bf 23.5} &
      \multicolumn{1}{c|}{\bf 31.6} &
      \multicolumn{1}{c|}{\bf 30.7} &
      \bf 39.8 \\ \hline

      
    \end{tabular}
    \label{Table1}
\end{table}
\section{Experiments}

We design and execute a series of synthetic and real-world experiments to both qualitatively and quantitatively evaluate our TRTM system with different cloths and tasks.

\begin{figure*}[t]
  \centering
  \includegraphics[width=0.8\paperwidth]{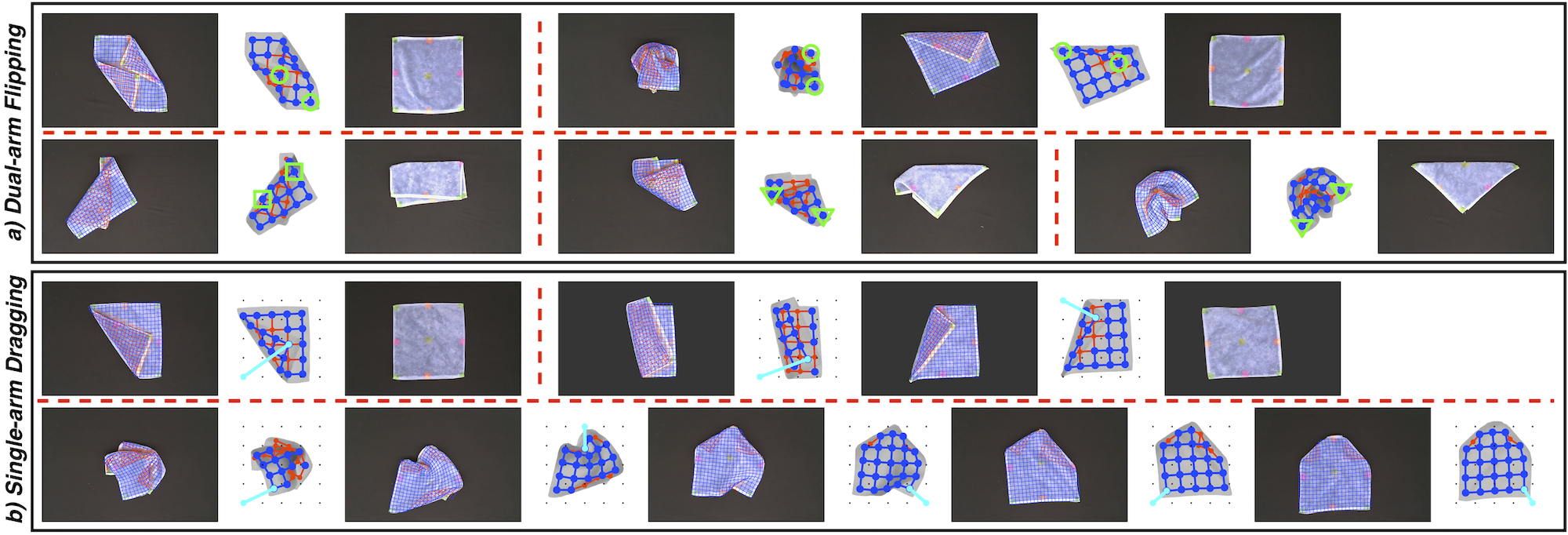}
  \vspace{-0.5em}
  \caption{{\bf Qualitative Evaluation of our Target-oriented Manipulation.} a) Dual-arm flipping by querying visible group vertex pairs according to different target configurations: flat, triangle, and rectangle. b) Single-arm flattening by sequentially dragging the visible group vertex to its canonical target position.}
  \label{Figure4}
\end{figure*}

\subsection{Single-view Cloth Reconstruction Results}

Three tiers of cloth configurations are generated within Blender by one-time dragging, two-times folding, and one-time dropping the synthetic template mesh $ T_{simu} $, which provides $ 120k $ cloth meshes and depth images for training.

To evaluate the real-world reconstruction, we randomly generate the above three cloth tiers using our marked template cloth $ T_{real} $, with a total of 600 cloth configurations. Since it is impossible to label the ground-truth mesh, we only label the marked keypoints, i.e., positions and visibilities of four corners, four middle-edges, and one center. We evaluate our reconstruction results using the supervised vertex-wise losses and pixel-wise losses, as well as the error of visible flags $ {L}_{vtx,f} / {L}_{key,f} $. The total loss $ {T}_{loss} $ is calculated using Equation (8) without $ {L}_{vtx,p} $ and $ {L}_{regu} $, as shown in Table \ref{Table1}.

The above reconstruction experiments demonstrate that our synthetic-trained GNN can explicitly and precisely reconstruct our template cloth both in simulation and the real world, with on average vertex-wise losses of 1.22 cm and 1.73 cm respectively. In addition, the direct reconstruction results of four other real-world cloths: one smaller square cloth $ {Sq}_{s} $, one larger square cloth $ {Sq}_{l} $, one rectangle cloth $ {Rect} $, and one shirt $ {shirt} $, are also demonstrated in Table \ref{Table1}.



\subsection{Dual-Arm Cloth Manipulation Results}

\begin{figure}[t]
  \centering
  \includegraphics[width=0.4\paperwidth]{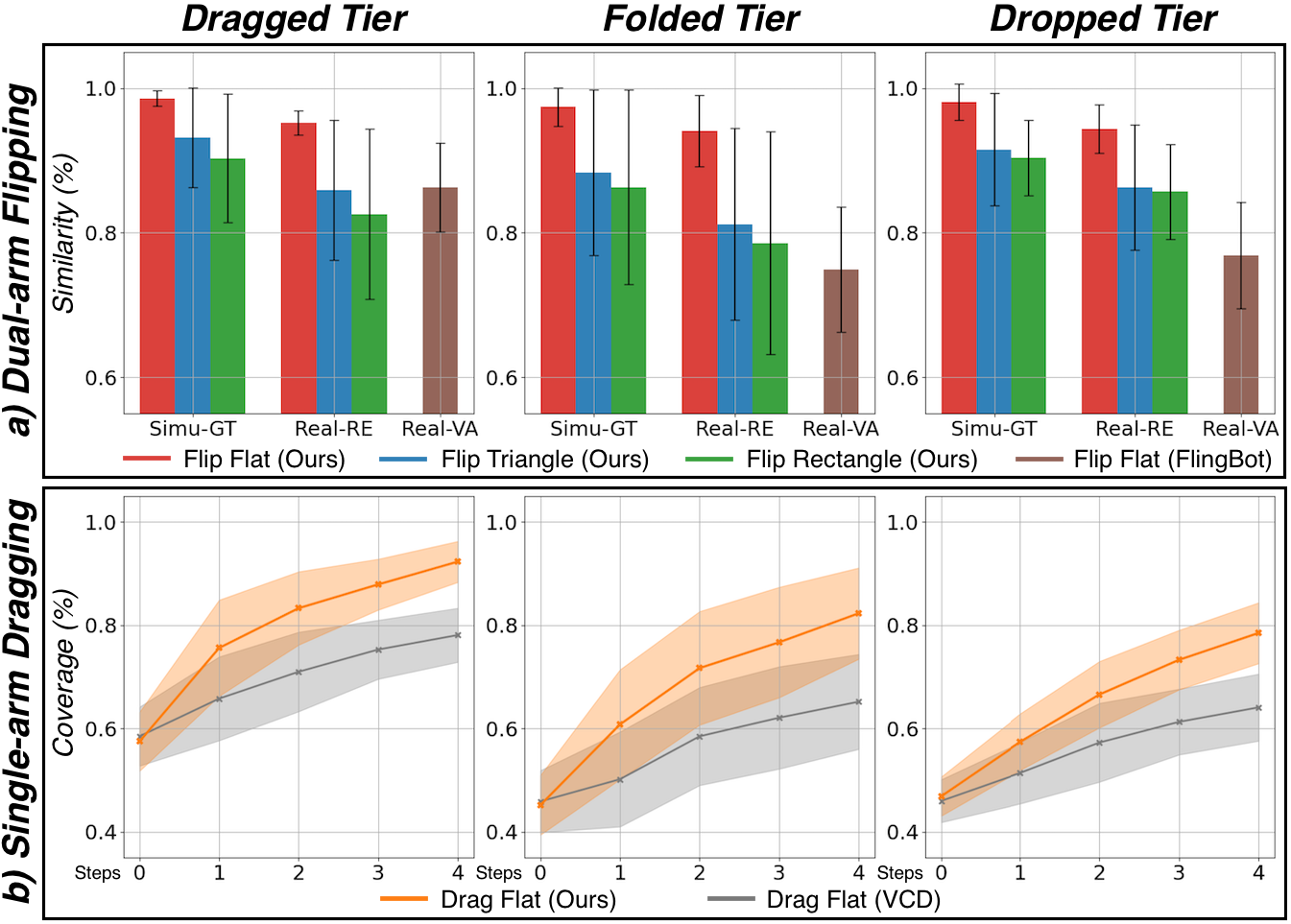}
  \caption{{\bf Quantitative Evaluation of our Target-oriented Manipulation.} a) Dual-arm flipping experiments with flat (red), triangle (blue), and rectangle (green) targets. We demonstrate the flipping results with the simulated ground truth meshes (Simu-GT), with the real-world reconstruction meshes (Real-GT), and value networks (Real-VA, flatten only). b) Real-world single-arm dragging for flattening experiments: ours (orange) and VCD (gray).}
  \label{Figure5}
\end{figure}

We dual-arm grasp-hang-and-flip the randomly dragged, folded, and dropped template cloths both in the simulation $ (3 \times 500 \times 3) $ and the real world $ (3 \times 50 \times 3) $, with three target configurations: flat, triangle, and rectangle, as shown in Fig. \ref{Figure4} (a) and Fig. \ref{Figure5} (a). In our work, the dual-arm manipulation episode is set as two for the flat target while only one for the triangle and rectangle targets. We evaluate the dual-arm flattened configurations using their top-view coverage values. For the triangle and rectangle targets, we evaluate through the top-view silhouette similarity between the flipped and the targeted configurations: $ 1 - L_{sil}(S_{flipped}, \; S_{targeted}) $.

We compare our real-world dual-arm flattening results with the FlingBot \cite{FlingBot}, where the flipping points are selected from the top-view color images using a task-specific value network trained within simulation with coverage rewards. For a fair comparison, we employ the FlingBot value network to select flipping points for $ 3 \times 50 $ randomly dragged, folded, and dropped template cloths, while keeping the rest setting the same, the results are shown in Fig. \ref{Figure5} (a), in brown.

Experimentally, using our explicit mesh representation, our dual-arm flipping agent can flip most of the randomly dragged, folded, and dropped cloths to flat ($ 97.6\% $ coverage) within two operation episodes, outperforming the implicit FlingBot agent ($ 78.3\% $ coverage). For triangle and rectangle targets, our dual-arm agent can achieve on average $ 84.5\% $ (Real-RE) / $ 90.8\% $ (Simu-GT) and $ 82.3\% $ (Real-RE) / $ 89.3\% $ (Simu-GT) top-view similarities within only one operation episode. Among three cloth tiers, the two-times folded cloths are mostly difficult to be reconstructed and manipulated.

\subsection{Single-arm Cloth Manipulation Results}

We modify one single-arm flattening strategy \cite{imitate} to our mesh group setting, where a canonical group target is assigned around the cloth image center, as shown in Fig. \ref{Figure4} (b). At each operation episode, our single-arm agent will drag the visible group vertex to its target position that has the longest distance. Within the real world, we single-arm flatten $ 3 \times 50 $ randomly dragged, folded, and dropped cloth configurations four times each, during which we report the top-view coverage values, as shown in Fig. \ref{Figure5} (b), in orange.

We compare our single-arm flattening results with the VCD work \cite{VCD}, where some random-shooting actions are optimized by a dynamic model trained within the simulation environment. For a fair comparison, we let the VCD agent generate actions to flatten $ 3 \times 50 $ randomly dragged, folded, and dropped template cloths, while keeping the rest setting the same, their results are shown in Fig. \ref{Figure5} (b), in gray.

\renewcommand{\arraystretch}{1.6}
\begin{table}[t]
    \centering
    \caption{Ablation Study of Template-based Reconstruction.}
    \begin{tabular}{|c||cccc|}
    \hline
    \multirow{2}{*}{Methods} & \multicolumn{4}{c|}{${T}_{real}$} \\ \cline{2-5} 
    & \multicolumn{1}{c|}{${L}_{key,f}$} & \multicolumn{1}{c|}{${L}_{key,p}$} & \multicolumn{1}{c|}{${L}_{sil}$} & ${L}_{cham}$ \\ \hline \hline
  
    
    \multicolumn{1}{|c||}{No attention weights} & \multicolumn{1}{c|}{16.6\%} & \multicolumn{1}{c|}{19.4mm} & \multicolumn{1}{c|}{10.4\%} & 11.6mm \\ \hline
  
    \multicolumn{1}{|c||}{No train augmentation} & \multicolumn{1}{c|}{22.1\%} & \multicolumn{1}{c|}{21.8mm} & \multicolumn{1}{c|}{11.6\%} & 12.5mm \\ \hline
    
    \multicolumn{1}{|c||}{No test augmentation} & \multicolumn{1}{c|}{17.4\%} & \multicolumn{1}{c|}{19.7mm} & \multicolumn{1}{c|}{10.7\%} & 11.3mm \\ \hline
  
    \multicolumn{1}{|c||}{{\bf Synthetic-trained GNN}} & \multicolumn{1}{c|}{{\bf 14.5\%}} & \multicolumn{1}{c|}{{\bf 17.3mm}} & \multicolumn{1}{c|}{{\bf 8.1\%}} & {\bf 9.6mm} \\ \hline \hline
  
  
    \multicolumn{1}{|c||}{Tune by training} & \multicolumn{1}{c|}{14.3\%} & \multicolumn{1}{c|}{17.0mm} & \multicolumn{1}{c|}{7.8\%} & 9.7mm \\ \hline
  
    \multicolumn{1}{|c||}{Tune by optimizing} & \multicolumn{1}{c|}{14.0\%} & \multicolumn{1}{c|}{16.6mm} & \multicolumn{1}{c|}{7.7\%} & 9.3mm \\ \hline
    \end{tabular}
    \label{Table2}
\end{table}

Experimentally, for those boundary-dragged cloth configurations, our single-arm flattening agent can directly recover the dragging action from the mesh distribution, and thus can unfold the cloth within two operation episodes. However, compared with the above dual-arm flipping agent, the single-arm dragging agent requires more operations to flatten a crumpled cloth, and usually sticks around some configurations where corners and edges are folded inside, like the last row in Fig. \ref{Figure4} (b). To flatten these states, some explicit reveal-and-drag actions can be introduced in future work.


\subsection{Ablation Studies}

{\bf Templated-based GNN.} In this section, we first examine some training and testing designs of our reconstruction model in front of the real-world template cloth $ T_{real} $, as shown in Table \ref{Table2}. The loss numbers represent the average reconstruction losses in front of the entire dragged, folded, and dropped real-world template cloth configurations.

We also employed some tuning strategies \cite{FlingBot, garmentdynamic} to our cloth reconstruction setting. Specifically, we additionally tune the synthetic GNN with part of the real-world depth observations (400) for another 50 epochs, supervised with the pixel-wise losses only. After this, we further optimize the entire reconstruction mesh with the pixel-wise losses only. The small improvements demonstrate that our synthetic trained model doesn't benefit much from the small-sized real-world data with only pixel-wise supervision.

\begin{figure}[t]
  \centering
  \includegraphics[width=0.4\paperwidth]{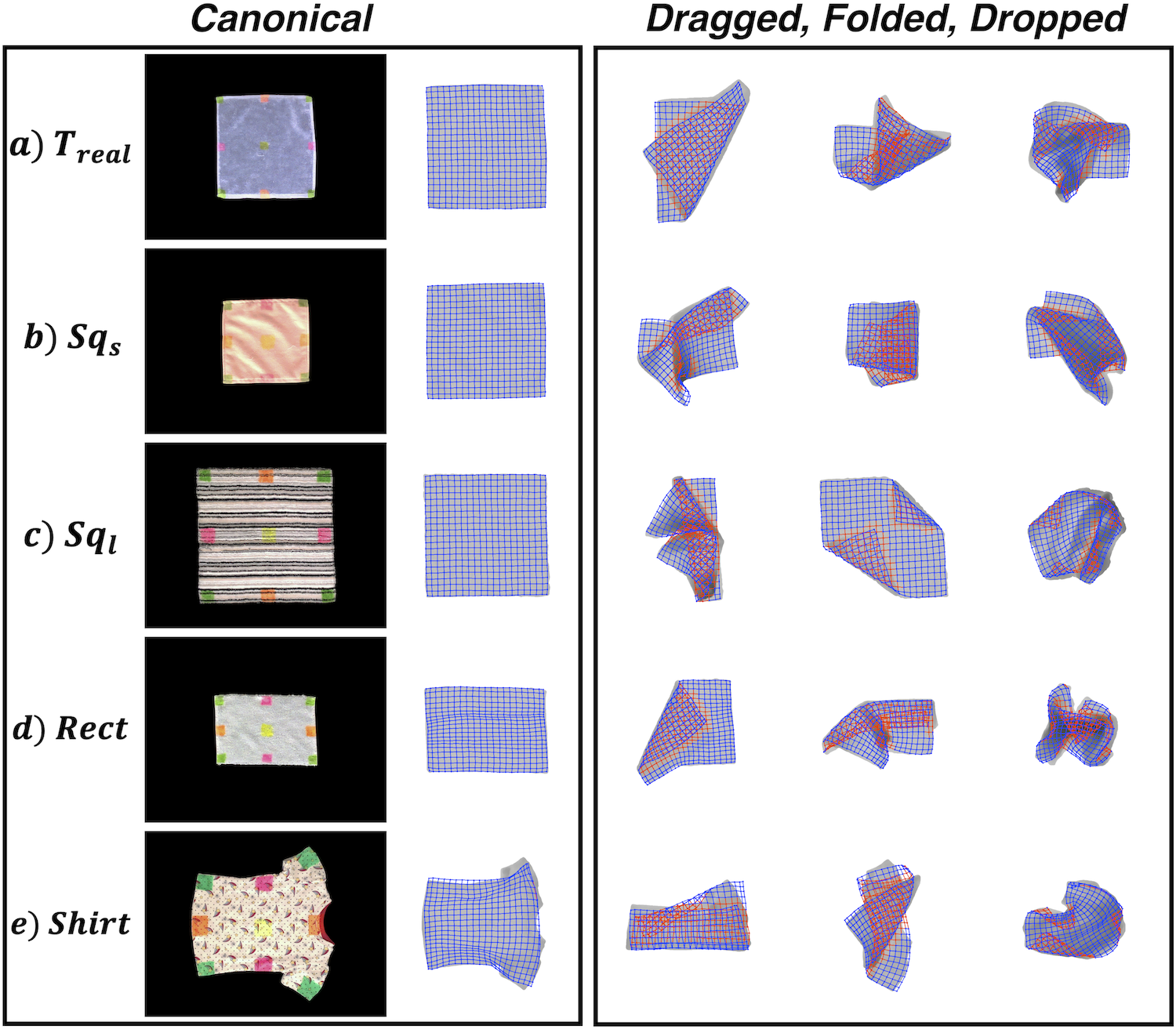}
  \caption{{\bf Qualitative Evaluation of our Template-based Reconstruction.} Direct reconstruction results of our synthetic-trained square-template GNN in front of different real-world cloths. From top to bottom: a) our sim-real registered $ 0.3m \times 0.3m $ template cloth $ {T}_{real} $; b) another $ 0.25m \times 0.25m $ stiffer square cloth $ {Sq}_{s} $; c) another $ 0.4m \times 0.4m $ softer square cloth $ {Sq}_{l} $; d) another $ 0.2m \times 0.3m $ rectangle cloth $ {Rect} $; e) another $ 0.3m \times 0.4m $ softer $ {Shirt} $ with two layers of rectangle bodies and sleeves. From left to right: canonical configurations; randomly dragged, folded, and dropped cloth configurations.}
  \label{Figure6}
\end{figure}

{\bf Sim-real Registration.} In this section, we demonstrate that our square-template GNN can be directly and reasonably applied to other daily cloths with a similar topology but different shapes, sizes, textures, and physical properties. To do so, we find another four real-world cloths $ ( {Sq}_{s}, \: {Sq}_{l}, \: {Rect}, \: {Shirt} ) $ and randomly generate their dragged, folded, and dropped cloth configurations with the size of 300 per cloth. We both quantitatively and qualitatively evaluate the direct reconstruction performance of our square-template GNN in front of the above real-world cloths, as shown in Table \ref{Table1} and Fig. \ref{Figure6}. More synthetic and real-world cloth data can be downloaded from our project website.

These experiments demonstrate that our synthetic-trained square-template GNN can be directly applied to daily square and rectangle cloths with different sizes, textures, and physical properties. It achieves nearly the same pixel-wise reconstruction results and is slightly vertex-wise less accurate. However, in front of those two-layered human garments that have totally different topologies from our template mesh, our square-template GNN can still fit a reasonable crumpled square mesh to their top-view depth observations, but reaching nearly doubled vertex-wise losses, especially when two garment layers are randomly detached and expanded. In future work, a synthetic shirt dataset can be simulated, from which a shirt GNN can be trained to achieve better reconstruction of crumpled shirts.

\section{Conclusion and Future Work}

In this paper, we propose a TRTM system that can precisely reconstruct and explicitly manipulate the randomly dragged, folded, and dropped cloths from their top-view observations only. Compared with the previous implicit and simplified cloth representations, our template-based reconstruction mesh explicitly indicates the positions and visibilities of the entire cloth mesh vertices. Experiments demonstrate that our explicit mesh representation promotes more explicit dual-arm and single-arm target-oriented manipulations, which can significantly outperform the previous implicit and task-specific cloth manipulation agents.

Regarding the future work, instead of fitting a TRTM system for each daily cloth with various template embeddings, we believe the fixed template used in our work can be further parameterized like the human body model \cite{SMPL}. From that stage, auto-template-registration \cite{garmentGNN, singleviewhuman} can be introduced to improve the reconstruction robustness in front of different cloths with various canonical properties.






\bibliographystyle{IEEEtran}
\bibliography{root.bib} 

\end{document}